\documentclass[letterpaper, 10 pt, conference]{ieeeconf}  
\usepackage{authblk}
\IEEEoverridecommandlockouts                              
                                                          
\newcommand \ignore[1]{}
\usepackage{booktabs}
\usepackage{graphicx}
\usepackage{adjustbox}
\usepackage{comment}
\usepackage{amsmath}
\usepackage{xcolor}
\definecolor{citeblue}{RGB}{0,114,178}  
\usepackage[colorlinks=true, citecolor=citeblue, linkcolor=citeblue, urlcolor=citeblue]{hyperref}

\title{\LARGE \bf
Beyond Attack Success Rate: A Multi-Metric Evaluation of Adversarial Transferability in Medical Imaging Models
}

\author{Emily Curl\textsuperscript{*}, Kofi Ampomah\textsuperscript{*}, Md Erfan\textsuperscript{*}, and Sayanton Dibbo\textsuperscript{$\dagger$}%
\thanks{\textsuperscript{*}These authors contributed equally to this work and are all considered first authors, and \textsuperscript{$\dagger$} is the corresponding author.}%
\thanks{The authors are with The University of Alabama, Tuscaloosa, USA. 
        Email: \{ecurl3, kampomah, merfan\}@crimson.ua.edu, sdibbo@ua.edu}%
}

\begin{document}

\maketitle
\thispagestyle{empty}
\pagestyle{empty}

\begin{abstract}

While deep learning systems are becoming increasingly prevalent in medical image analysis, their vulnerabilities to adversarial perturbations raise serious concerns for clinical deployment. These vulnerability evaluations largely rely on Attack Success Rate (ASR), a binary metric that indicates solely whether an attack is successful. However, the ASR metric does not account for other factors, such as perturbation strength, perceptual image quality, and cross-architecture attack transferability, and therefore, the interpretation is incomplete. This gap requires consideration, as complex, large-scale deep learning systems, including Vision Transformers (ViTs), are increasingly challenging the dominance of Convolutional Neural Networks (CNNs). These architectures learn differently, and it is unclear whether a single metric, e.g., ASR, can effectively capture adversarial behavior. To address this, we perform a systematic empirical study on four medical image datasets: PathMNIST, DermaMNIST, RetinaMNIST, and CheXpert. We evaluate seven models (VGG-16, ResNet-50, DenseNet-121, Inception-v3, DeiT, Swin Transformer, and ViT-B/16) against seven attack methods at five perturbation budgets, measuring ASR, Peak Signal-to-Noise Ratio (PSNR), Structural Similarity Index Measure (SSIM), and $L_2$ perturbation magnitude. Our findings show a consistent pattern: perceptual and distortion metrics are strongly associated with one another and exhibit minimal correlation with ASR. This applies to both CNNs and ViTs. The results demonstrate that ASR alone is an inadequate indicator of adversarial robustness and transferability. Consequently, we argue that a thorough assessment of adversarial risk in medical AI necessitates multi-metric frameworks that encompass not only the attack efficacy but also its methodology and associated overheads.

\end{abstract}

\section{Introduction}\label{sec:intro}

In the agentic AI era, machine learning (ML)- based systems, particularly deep neural networks (DNNs), have received significant attention and achieved exceptional performance across various domains and tasks. 
The tasks in different domains provides close to human accuracy in image recognition~\cite{lecun2002gradient, krizhevsky2012imagenet}, facial recognition~\cite{dibbo2024improving}, multimodal question answering~\cite{amebley2025neuro}, and highly accurate results in health monitoring~\cite{lien2024explaining}, speech recognition~\cite{hinton2012deep}, IoT security~\cite{vhaduri2021predicting}, natural language processing~\cite{andor2016globally}, biometric authentication~\cite{vhaduri2023bag}, wearable computing~\cite{sah2020adversarial}, and job automation~\cite{nasr2025scalable}. Major advances across domains have led to the use of deep learning in safety-critical applications, e.g., medical image analysis and clinical decision support. While DNNs are successful, they can still be attacked by adversaries. Szegedy et al.~\cite{szegedy2013intriguing} showed that even small, carefully planned perturbations/changes, often invisible to the human eye, can have an effect on model predictions by misleading the models to provide incorrect predictions. This vulnerability not only persists in high-stakes fields, such as medical image analysis, but also has serious consequences for patients. These adversarial examples are transferable and can fool a model, enabling \textit{black-box} attacks~\cite{liu2016delving}. Current studies thoroughly assess attack efficacy, focusing on optimizing ASRs, but often neglect a systematic analysis of the interrelationships among evaluation metrics across various models and architectures. ASRs only capture a binary idea of success~\cite{popovic2025debackdoor}, ignoring other potential factors, e.g., i) size of the perturbation, ii) quality of the perceptual image, and iii) overheads/cost of adversarial example generation. These limitations become challenging across heterogeneous architectures, e.g., Convolutional Neural Networks (CNNs) and Vision Transformers (ViTs), which fundamentally differ in feature representation and learning behavior, thereby introducing new uncertainties in adversarial transferability. ASR provides an incomplete and misleading picture of adversarial robustness, underscoring the importance of considering multiple metrics in evaluations. \textit Our objective is to assist researchers and clinicians in improving the reliability of medical AI systems and to inform the development of resilient multi-metric evaluation frameworks via an empirical investigation of the correlation between adversarial attack efficacy, perceptual image quality, and perturbations across CNNs and ViTs.

\textbf{Research Gap:} State-of-the-art research lacks a cohesive empirical framework to analyze the correlations between ASRs and additional metrics, including SSIM, PSNR~\cite{sara2019image}, and $L_2$ perturbations~\cite{bilgic2014fast}, in varying model architectures and a range of medical image datasets. It is still unclear whether adversarial attack success is inherently linked to perceptual distortion or perturbation strength, and how those associations vary across CNNs and ViTs.

\textbf{Our Contribution:} To address the gaps, we conduct a comprehensive empirical study of adversarial robustness and transferability across multiple CNNs and ViTs on different medical image datasets. Using Pearson~\cite{benesty2009pearson} and Spearman correlation analyses~\cite{sedgwick2014spearman}, we investigate the relationships between ASRs, perceptual quality (SSIM, PSNR), and perturbation magnitude $L_2$. Our findings show that while perceptual and distortion metrics are strongly interrelated, their relationship with ASRs remains consistently weak across architectures, highlighting a fundamental limitation of ASRs as a standalone metric and motivating the need for a multi-metric, relationship-aware evaluation framework.

\section{Background and Related Work}\label{sec:background_relatedwork}
We provide an overview of adversarial attacks, their transferability across architectures, and the limitations of existing evaluation metrics in medical image analysis contexts.
\subsection{Adversarial Attacks in Deep Learning}
Adversarial attacks exploit the sensitivity of DNNs to small, structured perturbations that cause misclassifications while remaining imperceptible to human observers. These attacks can range from perturbations to prompt injections~\cite{piet2024jatmo} or backdoors~\cite{peellawalage2026meta}. Goodfellow et al.~\cite{goodfellow2014explaining} attributed this vulnerability to the linear behavior of neural network computations in high-dimensional input spaces, introducing FGSM as a computationally efficient single-step attack. Iterative extensions such as I-FGSM ~\cite{kurakin2016adversarial} and Projected Gradient Descent (PGD) ~\cite{madry2017towards} expanded through repeated gradient updates while keeping perturbation budget within a fixed limit. Later approaches, Momentum-based variants such as MI-FGSM~\cite{dong2018boosting} and VMI-FGSM~\cite{wang2020enhancing} improve transferability by stabilizing gradient directions and reducing gradient variance. Even with this progress, ASR remains a primary metric for assessing how well an attack works.


\subsection{Transferability of Adversarial Examples}
A key property of adversarial examples is their transferability. Herein, inputs intended to fool one model can often mislead others, enabling black-box attacks~\cite{szegedy2013intriguing}. Techniques such as ensemble-based methods and input transformation have been shown to further improve this transferability~\cite{liu2016delving}, ~\cite{xie2019improving}. However, most existing studies focus on transferability within similar model families, particularly CNN-to-CNN settings. With the rise of ViTs~\cite{dosovitskiy2020image}, recent work has begun exploring cross-architecture transferability which are showing different robustness characteristics between convolutional and attention-based models. Despite these advances, there is a lack of understanding of how transferability interacts with evaluation metrics across heterogeneous architectures. 

\subsection{Medical Image Datasets and Domain Characteristics}
Medical image datasets present unique challenges for adversarial analysis due to their domain-specific properties. Standardized benchmarks such as MedMNIST v2~\cite{yang2023medmnist} provide lightweight and diverse datasets including PathMNIST (histopathology)~\cite{kather2019predicting}, DermaMNIST (dermatology)~\cite{tschandl2018ham10000}, and RetinaMNIST (ophthalmology)~\cite{yang2023medmnist}. In addition, large-scale clinical datasets such as CheXpert~\cite{irvin2019chexpert} provide real-world chest X-ray images annotated with multiple pathologies. Unlike natural image datasets, medical images often exhibit high structural similarity, subtle texture variations, and multi-label characteristics that significantly influence adversarial behavior. 

Small perturbations may alter diagnostically relevant features without introducing visible artifacts, and multi-label datasets such as CheXpert introduce complex decision boundaries where perturbations must simultaneously overcome the multiple classification thresholds~\cite{ma2021understanding}.


\subsection{Adversarial Robustness in Medical Image}
Finlayson et al.~\cite{finlayson2019adversarial} demonstrated that medical classifiers, including those for chest X-rays and dermatology images, are highly susceptible to adversarial manipulation, with perturbations capable of flipping diagnoses in ways that can directly impact patient outcomes. Subsequent work confirmed that even small perturbations can lead to clinically significant misclassifications across radiology, pathology, and ophthalmology~\cite{ma2021understanding}. Despite increasing attention, most existing work evaluates robustness primarily through ASR, without systematically exploring how dataset characteristics interact with adversarial effectiveness and perceptual degradation across heterogeneous architectures. 

\subsection{Evaluation Metrics and Research Gap}
We measured ASR as the percentage of inputs that were incorrectly classified after perturbation. As ASR shows a binary outcome, the metric doesn't say anything about the cost of perturbation or its effect on perception. Some existing image intensity metrics like PSNR~\cite{huynh2008scope}, SSIM~\cite{wang2004image}, and $L_2$ magnitude~\cite{carlini2017towards} provide information about the quality of perception and the strength of perturbation. The state-of-the-art research frequently employs these metrics in isolation, neglecting their interrelationships and their connection to ASR (Table~\ref{tab:metric_gap}).


\begin{table}[t]
\centering
\caption{Existing Evaluation Approaches and Limitations}
\label{tab:metric_gap}
\renewcommand{\arraystretch}{1.2}
\begin{tabular}{p{3.0cm}p{3.5cm}}
\toprule
\textbf{Claim} & \textbf{Supported By} \\
\midrule
ASR-focused optimization & Madry et al.~\cite{madry2017towards}, Carlini \& Wagner~\cite{carlini2017towards} \\
Transferability via ASR & Liu et al.~\cite{liu2016delving}, Dong et al.~\cite{dong2018boosting} \\
Perceptual metrics independently & Laidlaw et al.~\cite{laidlaw2020perceptual}, Wang et al.~\cite{wang2004image} \\
No unified framework & Croce et al.~\cite{croce2020robustbench} \\
\bottomrule
\end{tabular}
\end{table}

\section{Methodology}
This section discusses an experimental pipeline for generating adversarial examples on surrogate CNN and ViT models. We also evaluate their transferability across image datasets with different perturbations using multiple performance and perceptual metrics.
\subsection{Overview}
\begin{figure}[!htbp]
\centering
\includegraphics[width=0.41\textwidth]{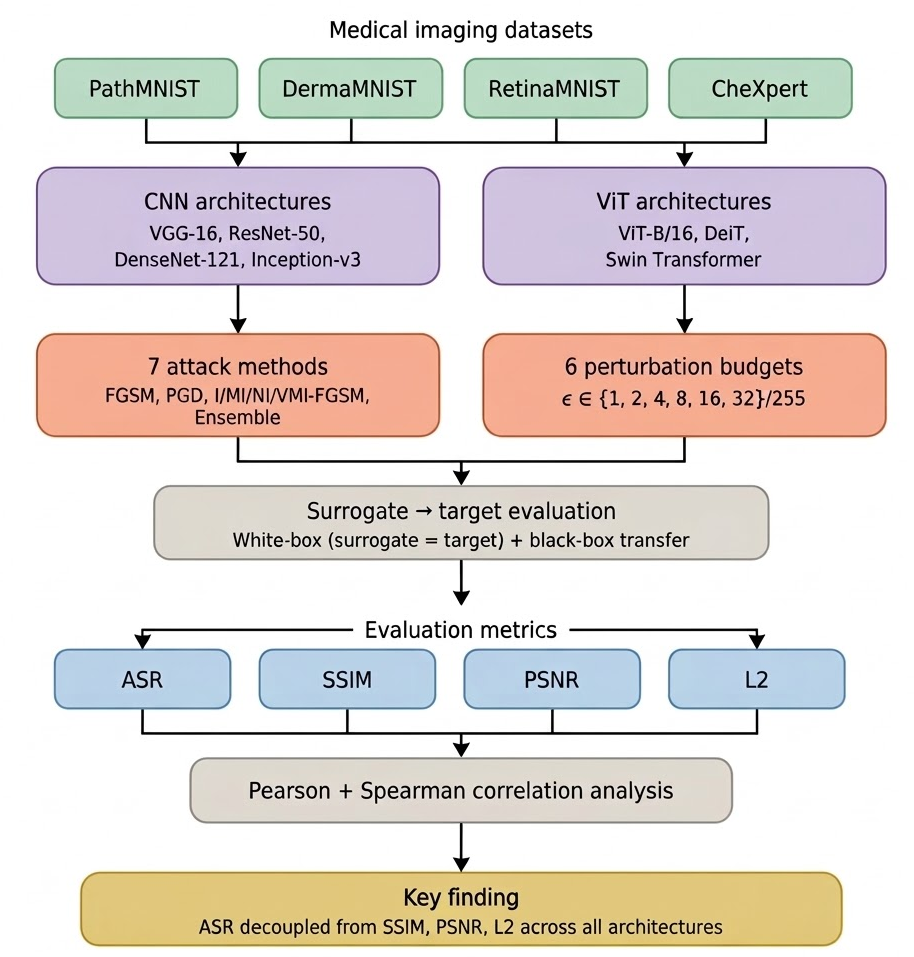}
\caption{Overview of the experimental pipeline. Four medical image datasets are evaluated across CNN and ViT families using seven attacks under five $\epsilon$ budgets, yielding 3,500 configurations. Correlation analysis shows that ASR is decoupled from perceptual/distortion metrics.}
\label{fig:overview}
\end{figure}

This study investigates adversarial transferability across heterogeneous deep learning architectures, specifically CNNs and ViTs, in medical image. A unified experimental pipeline (Figure~\ref{fig:overview}) is designed to ensure consistent evaluation across model families, enabling controlled comparison of adversarial behavior under identical conditions. For each model family, adversarial examples are generated using a surrogate model and evaluated on all models within the same family, producing both white-box (surrogate = target) and black-box (surrogate $\neq$ target) results.

\subsection{Datasets}
We utilize four publicly available medical image datasets spanning multiple clinical domains (Table~\ref{tab:dataset_overview}): PathMNIST (histopathology, 9 classes), DermaMNIST (dermatology, 7 classes), RetinaMNIST (ophthalmology, 5 classes), and CheXpert (chest X-ray radiology, 14 classes). The first three are drawn from the MedMNIST v2 benchmark and represent single-label classification tasks, while CheXpert involves multi-label classification where multiple pathologies may coexist within a single image. All images are resized to $224 \times 224$ pixels to ensure compatibility across architectures. Grayscale images (CheXpert, RetinaMNIST) are converted to three-channel RGB representations, and all inputs are normalized to the range $[-1,1]$.
\begin{table}[t]
\centering
\caption{Overview of Medical Image Datasets}
\label{tab:dataset_overview}
\renewcommand{\arraystretch}{1.2}
\resizebox{\columnwidth}{!}{%
\begin{tabular}{llcp{3.8cm}}
\toprule
\textbf{Dataset} & \textbf{Modality} & \textbf{\#Cls} & \textbf{Key Characteristics} \\
\midrule
DermaMNIST~\cite{tschandl2018ham10000} & Dermatoscopy & 7 & Skin lesion; high inter-class similarity \\
PathMNIST~\cite{kather2019predicting} & Histopathology & 9 & Tissue classification; high texture variation \\
RetinaMNIST~\cite{yang2023medmnist} & OCT & 5 & Retinal disease; subtle structural differences \\
CheXpert~\cite{irvin2019chexpert} & Chest X-ray & 14 & Multi-label clinical; distribution shifts \\
\bottomrule
\end{tabular}%
}
\end{table}

\subsection{Model Architectures}
We evaluate two families of models representing the dominant paradigms in medical image classification. The CNN family includes VGG-16~\cite{simonyan2014very}, ResNet50~\cite{he2016deep}, DenseNet121~\cite{huang2017densely}, and Inception-v3~\cite{szegedy2016rethinking}, which share a common inductive bias toward local feature extraction through convolutional operations despite differing in depth and architectural design. The transformer family includes ViT-B/16~\cite{dosovitskiy2020image}, DeiT~\cite{touvron2021training}, and Swin-Transformer~\cite{liu2021swin}, which capture long-range dependencies through self-attention, representing a fundamentally different approach to feature representation. All models are initialized with ImageNet-pretrained weights and evaluated in inference mode without fine-tuning, isolating architectural differences from training-specific effects.

\subsection{Adversarial Attacks}

We assessed seven adversarial attack methods, ranging from single-step to iterative gradient-based techniques, with increasing complexity. The FGSM~\cite{goodfellow2014explaining} is a single-step baseline that uses perturbation inputs along the loss gradient's sign in one update. Moreover, PGD~\cite{dibbo2024lcanets++} and I-FGSM~\cite{lad2024fast} build on FGSM by making iterative gradient updates with projection. The techniques make adversarial examples stronger within the perturbation budget. Also, MI-FGSM and NI-FGSM use momentum and Nestervov acceleration, respectively, to keep gradient directions stable across iterations. However, VMI-FGSM also makes transferability better by using neighborhood sampling to reduce the variance in gradient estimation. Finally, the ensemble attack combines gradients from all surrogate models in the same architectural family. Since ensemble techniques reduce surrogate overfitting, we find them effective for transfer.

\subsection{Perturbation Settings}
Adversarial perturbations are generated under five $L_\infty$-bounded budgets: $\{2/255, 4/255, 8/255, 16/255, 32/255\}$. This range spans from near-imperceptible perturbations ($\epsilon = 2/255$) to clearly visible modifications ($\epsilon = 32/255$), enabling systematic analysis of the trade-off between perturbation strength and adversarial effectiveness across the full spectrum of attack intensities.

\subsection{Evaluation Metrics}
We evaluate adversarial performance using five complementary metrics. ASR measures the proportion of inputs successfully misclassified after perturbation, while classification accuracy captures residual model performance under attack. To assess the perceptual impact of perturbations, we use SSIM, which quantifies structural and perceptual similarity between clean and adversarial inputs, and PSNR, which measures the peak signal-to-noise ratio. $L_2$ perturbation magnitude captures the Euclidean distance between clean and adversarial inputs, providing a direct measure of perturbation strength. Together, ASR and ACC reflect classification outcome, while SSIM, PSNR, and $L_2$ characterize the nature and cost of the perturbation itself. 

\subsection{Threat Model}
We adopt an untargeted adversarial attack setting under an $L_\infty$ norm constraint. The adversary has full white-box access to a surrogate model, including its architecture, parameters, and gradients, but has no access to the target model. This reflects a realistic black-box transfer attack scenario commonly encountered in practice, where adversarial examples crafted using a proxy model are deployed against an unknown target system. Formally, given an input $x$ with label $y$, the adversarial example $x_{\text{adv}}$ satisfies $\|x_{\text{adv}} - x\|_{\infty} \leq \epsilon$, and the objective is to maximize misclassification while keeping perturbations within the specified bound.

\section{Experiments}
In this section, we conduct large-scale adversarial attack experiments across multiple models, datasets, and settings. We then evaluate and analyze the results using diverse metrics and correlation studies to understand their relationships.
\subsection{Experimental Setup}
Experiments are conducted across all combinations of datasets, perturbation budgets, surrogate models, target models, and attack algorithms, yielding a total of 3,500 experimental configurations. The ViT family contributes 1,260 configurations (4 datasets $\times$ 5 $\epsilon$ values $\times$ 3 surrogates $\times$ 3 targets $\times$ 7 attacks, with 3 white-box on the diagonal), while the CNN family contributes 2,240 configurations (4 datasets $\times$ 5 $\epsilon$ values $\times$ 4 surrogates $\times$ 4 targets $\times$ 7 attacks). For each configuration, adversarial examples are generated using the surrogate model and then evaluated on all target models within the same architecture family.

\subsection{Evaluation Protocol}
For each (Dataset $\times$ $\epsilon$ $\times$ Surrogate $\times$ Attack) configuration, adversarial examples are generated for the entire test set. These perturbed samples are then forwarded through each target model to compute ASR and ACC. Perceptual and perturbation metrics (SSIM, PSNR, $L_2$) are computed once per attack configuration, as they depend only on the difference between clean and adversarial inputs and are independent of the target model. This separation ensures that perceptual metrics reflect the properties of the perturbation itself, while ASR captures the downstream classification effect on each target.

\subsection{Correlation Analysis}
We calculate the Pearson and Spearman correlation coefficients for 3,500 experimental configurations to examine the relationship among metrics. Pearson identifies linear correlations between metric pairs, while Spearman identifies monotonic rank-order correlations. The relationship ensures that the analysis remains resilient to non-linear dependencies that may emerge in various attack scenarios and perturbation budgets. In our research, we calculated the correlations separately for CNNs, ViTs, and the combined dataset.

\subsection{Implementation Details}
We used Pytorch \textit{TIMM} library to load models and make predictions for our experiments. We used NVIDIA GPU hardware, where models were run with a batch size of 32. We used a mean and standard deviation of 0.5 per channel to normalize the input images. We use $K = 10$ iterations and a momentum coefficient of $\mu = 1.0$ for MI-FGSM and NI-FGSM for all iterative attack methods. For VMI-FGSM, we employed neighborhood sampling with $N = 20$ samples for variance tuning, in alignment with the original formulation ~\cite{wang2020enhancing}. 


\section{Results}\label{sec:results}
This section presents the experimental findings, showing how adversarial attacks behave across models and datasets. 
\subsection{Dataset-Level Adversarial Effectiveness}

\begin{figure}[!htbp]
\centering
\includegraphics[width=\columnwidth]{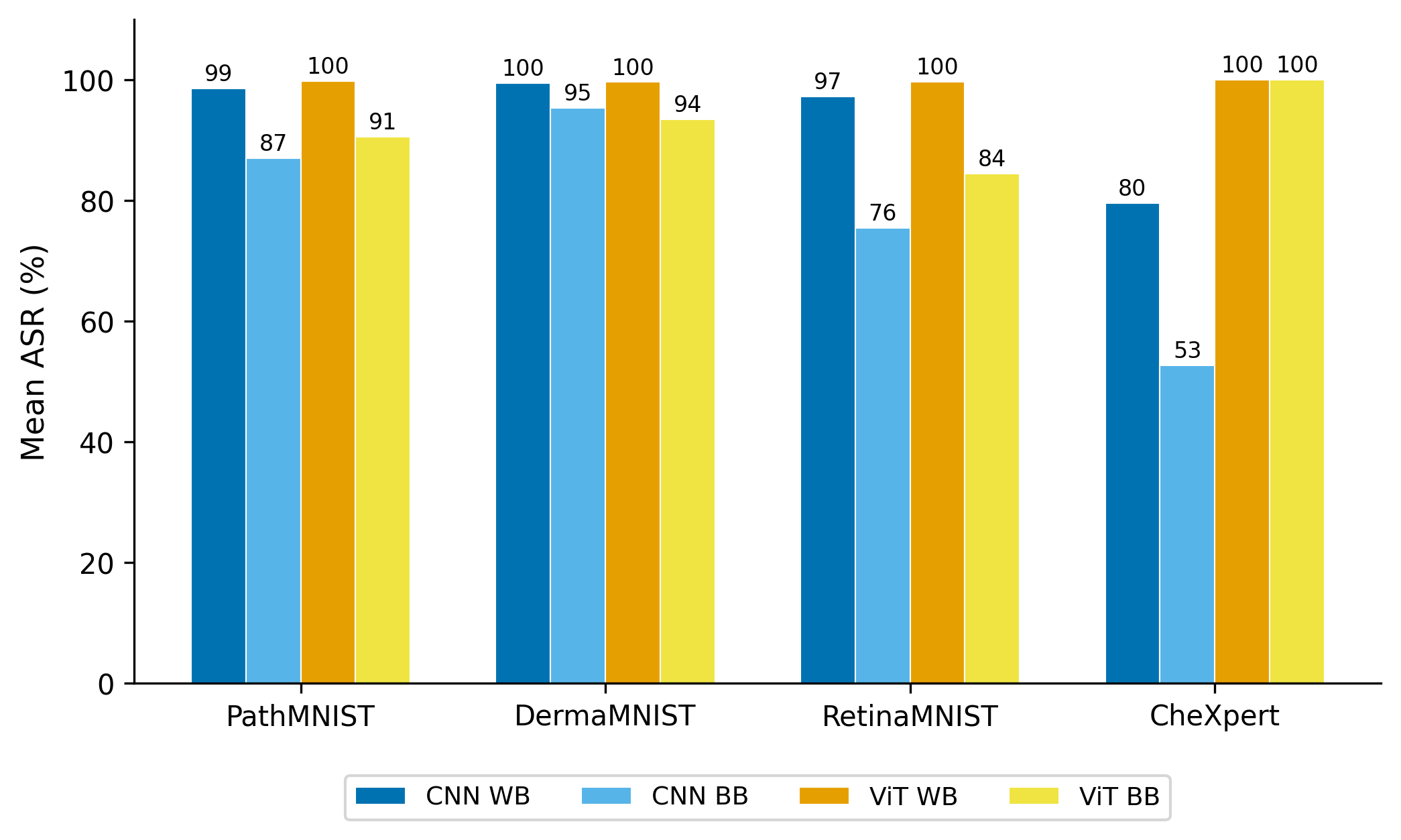}
\caption{Mean ASR by dataset, architecture, and attack setting. CheXpert CNN black-box (52.8\%) is notably lower than all other configurations.}
\label{fig:per_dataset_asr}
\end{figure}

We conducted tests on various datasets, architectures, and attack configurations. The average ASR values for each dataset, architecture, and attack configuration (white-box versus black-box) are depicted in Figure~\ref{fig:per_dataset_asr}. In our results, the ViT models achieve superior ASR values than CNN models across all datasets, regardless of whether the context is white-box or black-box. Interestingly, ViT white-box attacks achieve almost optimal ASR (99.5\% or above) across all datasets, whereas ViT black-box transfers remain elevated (from 84.5\% to 100.0\%). On the other hand, the CNN white-box ASR ranges from 79.6\% (CheXpert) to 99.5\% (DermaMNIST), while the CNN black-box ASR ranges from 52.8\% (CheXpert) to 95.4\% (DermaMNIST). The CheXpert CNN anomaly is noteworthy: the CNN black-box ASR on CheXpert (52.8\%) is significantly lower than on any other dataset, yet the ViT models achieve 100\% ASR on CheXpert in both configurations. The discrepancy is likely due to CheXpert employing a multi-label categorization framework, which complicates CNN-to-CNN adversarial transmission but does not impede ViT-to-ViT transfer.

\subsection{Attack Method Comparison}

\begin{figure}[!htbp]
\centering
\includegraphics[width=\columnwidth]{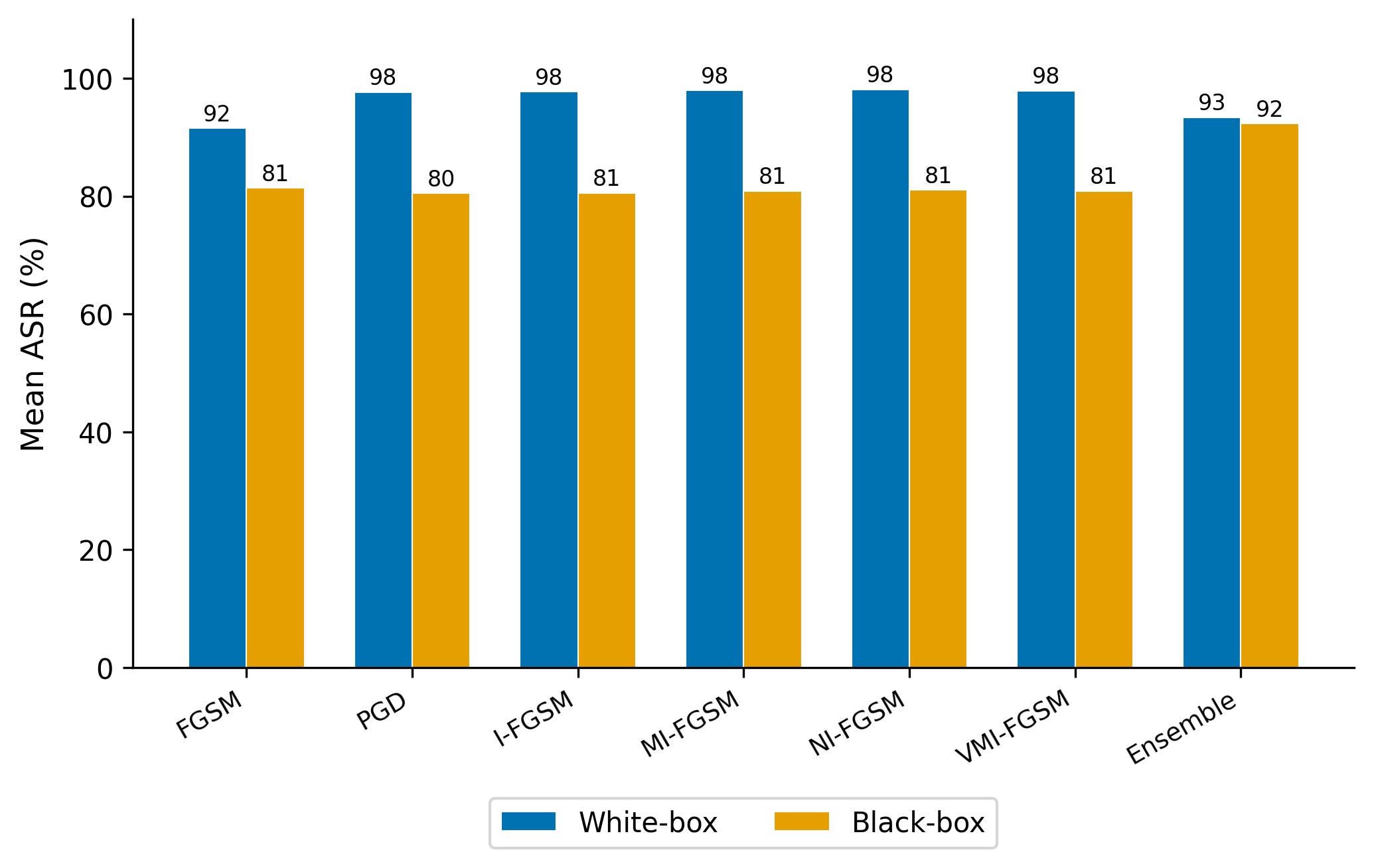}
\caption{Mean ASR by attack method. Ensemble has the smallest WB/BB gap (93.4\% vs.\ 92.3\%).}
\label{fig:per_attack_asr}
\end{figure}

Figure~\ref{fig:per_attack_asr} compares the effectiveness of the seven attack methods across all models and datasets.

In the white-box setting, iterative attacks (PGD, I-FGSM, MI-FGSM, NI-FGSM, VMI-FGSM) achieve consistently high ASR ($\geq$97.7\%), while single-step FGSM is slightly lower at 91.5\%. In the black-box setting, all individual attacks converge to a narrow band of 80.5-81.4\%, with the notable exception of the Ensemble attack, which achieves 92.3\% black-box ASR. This makes the Ensemble attack's gap the smallest of all methods (93.4\% vs. 92.3\%, a gap of only 1.1 percentage points), indicating that gradient aggregation across multiple surrogates is an effective strategy for generating transferable adversarial samples.

\subsection{Adversarial Transferability Across Models}

\begin{figure}[!htbp]
\centering
\includegraphics[width=0.47\textwidth]{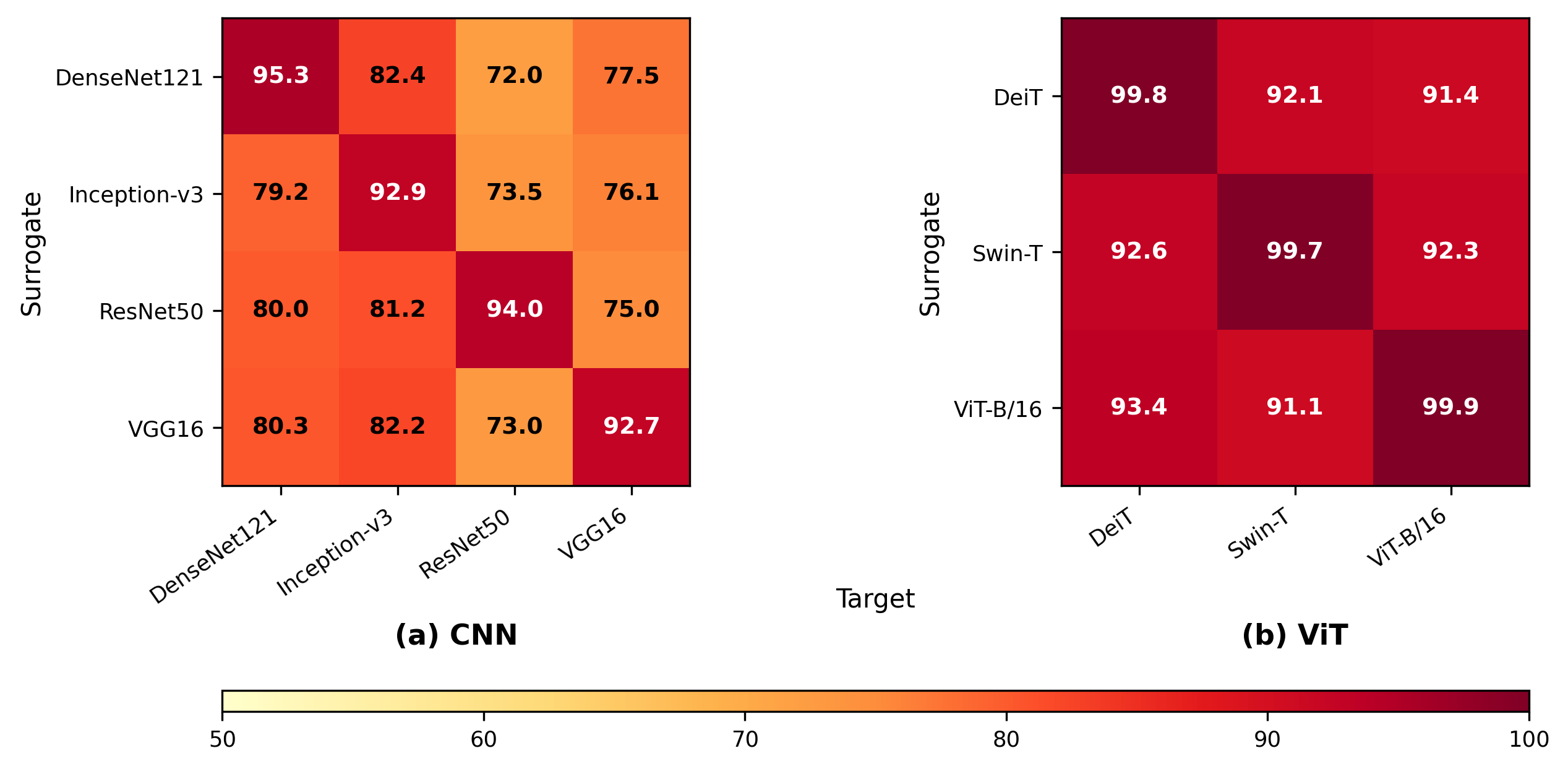}
\caption{Transferability matrices (mean ASR \%). (a)~CNN: structured, asymmetric (ResNet50 hardest target). (b)~ViT: uniformly high (91.1--93.4\%).}
\label{fig:transfer_matrices}
\end{figure}

To examine pairwise transferability patterns, we compute the mean ASR for each surrogate-target pair within each architecture family. Figure~\ref{fig:transfer_matrices} presents the resulting transferability matrices as heatmaps.

Within the CNN family, transferability exhibits a clear asymmetric structure. Inception-v3 is consistently the easiest target to attack, receiving 81.2-82.4\% black-box ASR across all surrogates, while ResNet50 proves the most resistant, receiving only 72.0-73.5\% black-box ASR. This asymmetry indicates that CNN architectures, while possessing a common convolutional inductive bias, acquire sufficiently unique feature representations that adversarial perturbations do not disseminate uniformly across all pairs. 
The ViT family, on the other hand, has a very consistent transferability, with all black-box surrogate-target pairs getting 91.1–93.4\% of  ASR and not much difference between pairs. This consistency suggests that attention-based architectures typically converge towards more similar adversarial vulnerability patterns in comparison to their convolutional counterparts. The practical implication for medical image analysis is significant: in ViT-based implementations, an adversary with access to a single ViT model may establish attacks that consistently propagate to other ViT models, regardless of the targeted architecture.


\subsection{Effect of Perturbation Budget}

\begin{figure}[!htbp]
\centering
\includegraphics[width=\columnwidth]{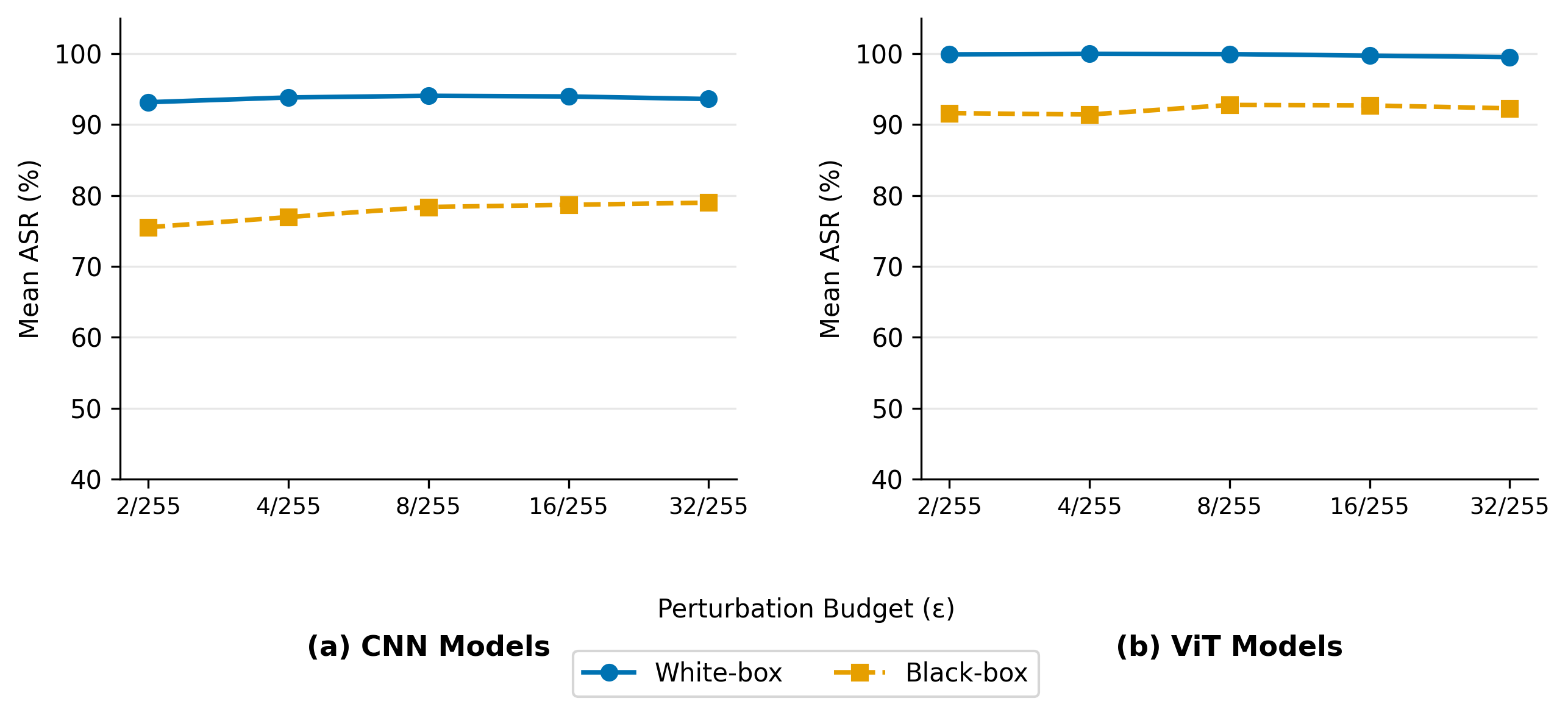}
\caption{ASR vs.\ $\epsilon$. ASR remains flat across the entire range, supporting the decoupling between perturbation magnitude and adversarial success.}
\label{fig:asr_vs_eps}
\end{figure}

The average ASR about the perturbation budget $\epsilon$ for both architectural families in white-box and black-box settings is depicted in Figure~\ref{fig:asr_vs_eps}. We can observe how little the ASR curves change across the full range of $\epsilon$. Among CNN models, white-box ASR remains between 93.1\% and 94.1\%, and black-box ASR between 75.5\% and 79.0\%. The pattern is even more pronounced for ViTs, where white-box ASR remains within 99.5-100.0\% and black-box ASR within 91.4-92.8\%. We found the perturbation magnitude increases 16-fold from $2/255$ to $32/255$, yet ASR barely shifts. The finding reinforces the decoupling, as we observed earlier: pushing $\epsilon$ higher does not translate into meaningfully greater adversarial success. The results suggest that ASR and perturbation magnitude capture quite different dimensions of adversarial behavior.

\subsection{Correlation Analysis: ViT Models}

To investigate the relationships between adversarial evaluation metrics for transformer-based architectures, we compute both Pearson and Spearman correlation coefficients across all 1,260 ViT configurations. The results are summarized in Table~\ref{tab:vit_corr}.

\begin{table}[t]
\centering
\caption{Correlation Analysis for ViT Models}
\label{tab:vit_corr}
\begin{tabular}{lcc}
\toprule
\textbf{Metric Pair} & \textbf{Pearson ($r$)} & \textbf{Spearman ($\rho$)} \\
\midrule
SSIM vs PSNR & 0.8766 & 0.9739 \\
SSIM vs $L_2$ & $-$0.9596 & $-$0.9739 \\
PSNR vs $L_2$ & $-$0.9186 & $-$1.0000 \\
ASR vs SSIM & 0.0479 & 0.1054 \\
ASR vs PSNR & $-$0.0338 & 0.0175 \\
ASR vs $L_2$ & 0.0139 & $-$0.0178 \\
\bottomrule
\end{tabular}%
\end{table}

SSIM and PSNR exhibit a strong positive correlation (Pearson: 0.88, Spearman: 0.97), indicating consistent agreement between perceptual quality metrics. Both show strong negative correlations with $L_2$, confirming that larger perturbations degrade perceptual quality. The relationship between PSNR and $L_2$ is nearly perfectly inverse (Spearman: $-$1.0), reflecting a consistent monotonic relationship between noise magnitude and signal quality. In contrast, ASR demonstrates near-zero correlations with all three metrics ($|r| \leq 0.05$, $|\rho| \leq 0.11$), indicating that adversarial success in ViT models is largely independent of both perceptual quality and perturbation magnitude.

\subsection{Correlation Analysis: CNN Models}

We repeat the correlation analysis across all 2,240 CNN configurations. The results, summarized in Table~\ref{tab:cnn_corr}, show a nearly identical structural pattern to the ViT analysis.

\begin{table}[t]
\centering
\caption{Correlation Analysis for CNN Models}
\label{tab:cnn_corr}
\begin{tabular}{lcc}
\toprule
\textbf{Metric Pair} & \textbf{Pearson ($r$)} & \textbf{Spearman ($\rho$)} \\
\midrule
SSIM vs PSNR & 0.8730 & 0.9743 \\
SSIM vs $L_2$ & $-$0.9595 & $-$0.9743 \\
PSNR vs $L_2$ & $-$0.9166 & $-$1.0000 \\
ASR vs SSIM & $-$0.1657 & $-$0.1505 \\
ASR vs PSNR & $-$0.0461 & $-$0.0409 \\
ASR vs $L_2$ & 0.0333 & 0.0409 \\
\bottomrule
\end{tabular}%
\end{table}

Perceptual and distortion-based metrics maintain strong internal relationships (SSIM--PSNR Pearson: 0.87; PSNR--$L_2$ Spearman: $-$1.0), while ASR correlations with all three metrics remain weak ($|r| \leq 0.17$, $|\rho| \leq 0.15$). The ASR-SSIM correlation in CNNs ($r = -0.17$) is a little stronger than in ViTs ($r = 0.05$), but it is still well below any threshold of practical significance. This consistency across architectures strengthens the notion that ASR decoupling is not merely an artifact of a particular model family, but rather a fundamental characteristic of the relationship between these metrics and adversarial behavior.

\subsection{Overall Correlation Analysis (CNN + ViT)}
\begin{figure}[!htbp]
\centering
\includegraphics[width=0.45\textwidth]{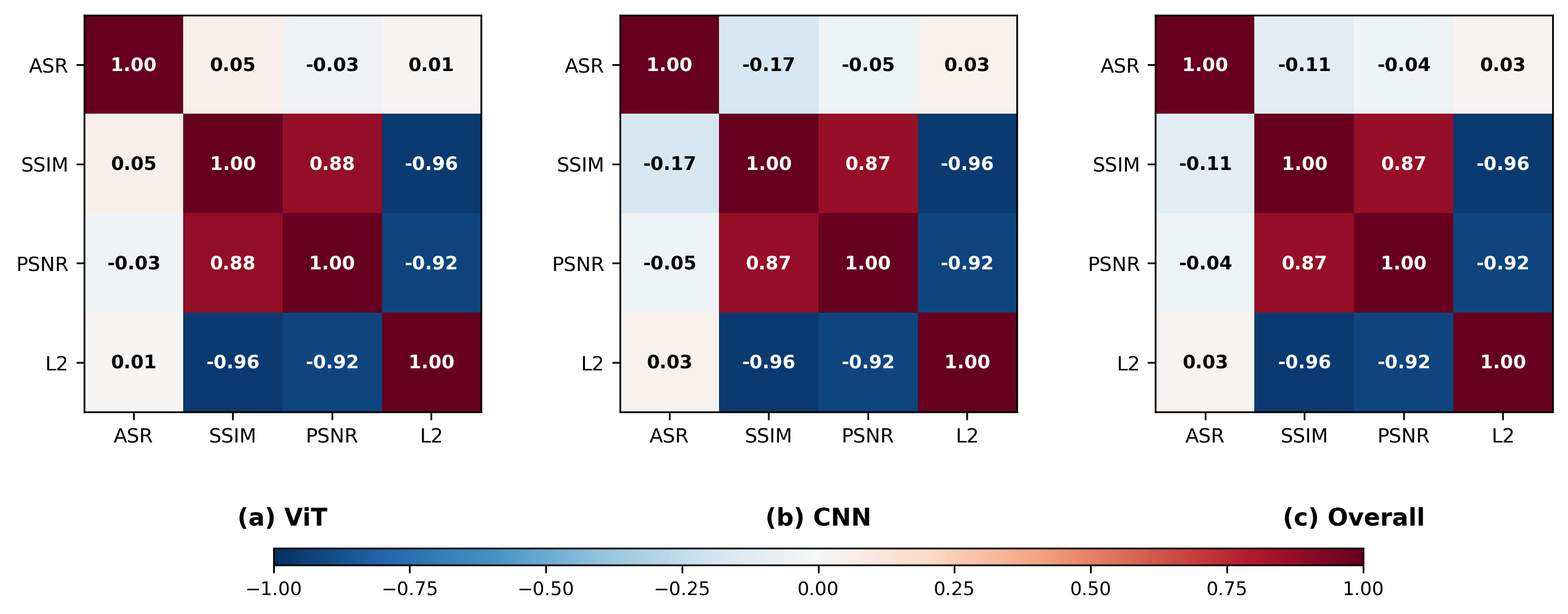}
\caption{Pearson (lower triangle) / Spearman (upper triangle) correlation matrices for (a)~ViT, (b)~CNN, and (c)~overall. ASR row/column is consistently pale (near-zero), while the SSIM--PSNR--$L_2$ block shows strong interrelationships.}
\label{fig:corr_heatmaps}
\end{figure}

\begin{figure}[!htbp]
\centering
\includegraphics[width=0.45\textwidth]{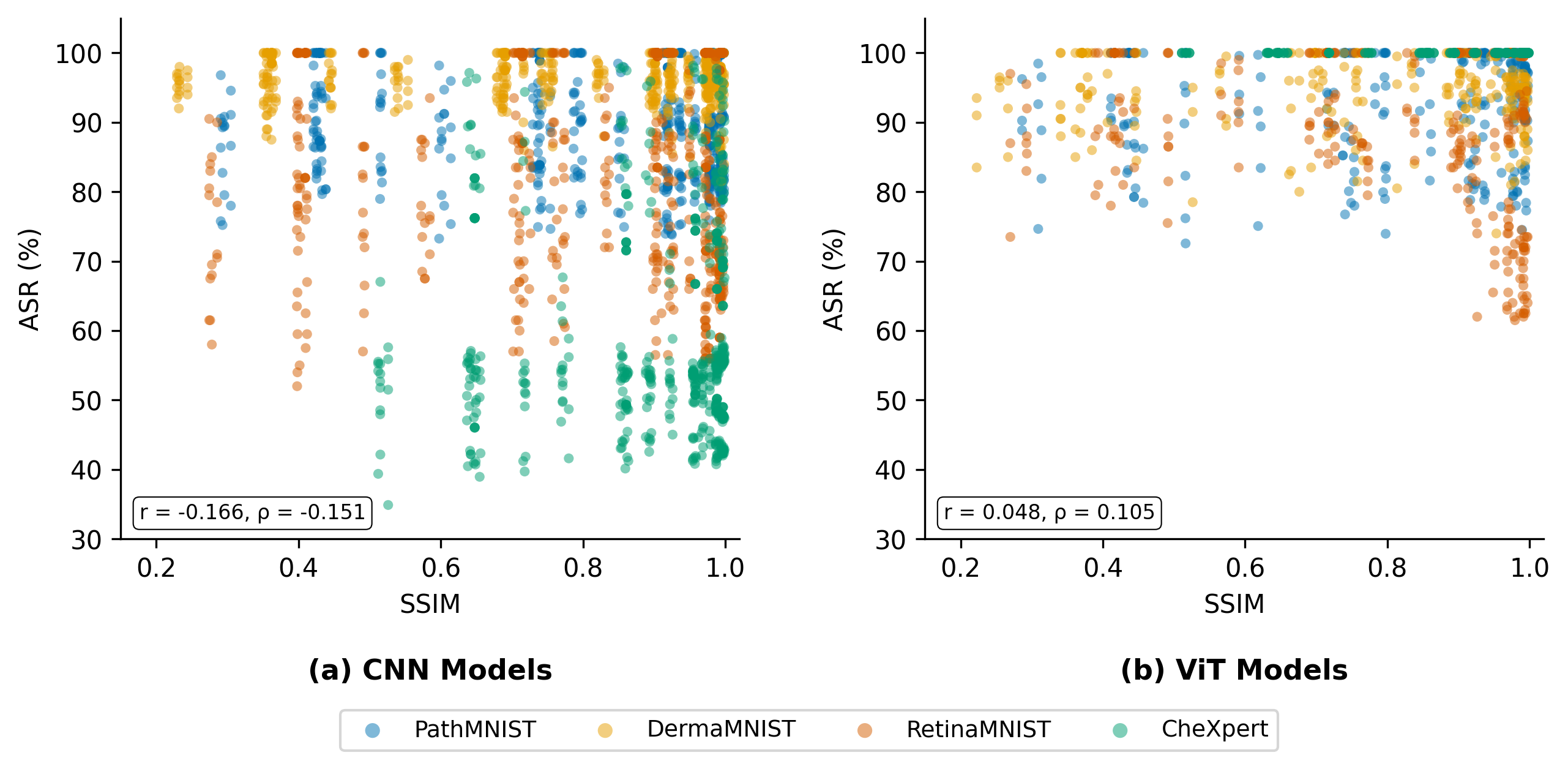}
\caption{ASR vs.\ SSIM for (a)~CNN and (b)~ViT models, colored by dataset. No discernible trend confirms the near-zero correlation between adversarial success and perceptual quality.}
\label{fig:asr_vs_ssim}
\end{figure}

We combined all 3,500 configurations and show the overall correlation matrix in Table~\ref{tab:overall_corr}. We present the Pearson and Spearman heat maps for ViT, CNN, and combined analyses side by side in Figure~\ref{fig:corr_heatmaps}, while Figure~\ref{fig:asr_vs_ssim} visualizes the ASR-SSIM relationship through scatter plots across all datasets and architectures. The overall analysis shows two key findings across both model families

\begin{itemize}
    \item \textbf{Metric Coupling:} We found that perceptual and distortion-based metrics form a tightly coupled cluster. Specifically, SSIM, PSNR, and $L_2$ are strongly interrelated ($|r| \geq 0.87$, $|\rho| \geq 0.97$). This finding reflects the intuitive relationship between perturbation magnitude and perceptual degradation.
    
    \item \textbf{ASR Decoupling:} ASR is consistently decoupled from this cluster, with all ASR correlations remaining below $|r| = 0.11$ and $|\rho| = 0.07$, regardless of architecture or dataset. As shown in the scatter plots in Figure~\ref{fig:asr_vs_ssim}, the decoupling is visually apparent; data points spread broadly across the ASR axis at every SSIM value, showing no discernible trend in either CNN or ViT models.
\end{itemize}


\begin{table}[t]
\centering
\caption{Overall Correlation Analysis (CNN + ViT)}
\label{tab:overall_corr}
\begin{tabular}{lcc}
\toprule
\textbf{Metric Pair} & \textbf{Pearson ($r$)} & \textbf{Spearman ($\rho$)} \\
\midrule
SSIM vs PSNR & 0.8742 & 0.9750 \\
SSIM vs $L_2$ & $-$0.9595 & $-$0.9750 \\
PSNR vs $L_2$ & $-$0.9173 & $-$1.0000 \\
ASR vs SSIM & $-$0.1076 & $-$0.0646 \\
ASR vs PSNR & $-$0.0422 & $-$0.0328 \\
ASR vs $L_2$ & 0.0272 & 0.0329 \\
\bottomrule
\end{tabular}%
\end{table}

We found that the pearson correlations between ASR and SSIM tend to zero across the board: CheXpert ($-$0.001), DermaMNIST (0.038), PathMNIST (0.001), and RetinaMNIST ($-$0.125). We found the RetinaMNIST as the strongest, and even that sits at just $r = -0.125$. The finding shows a pattern that holds in histopathology, dermatology, ophthalmology, and radiology. The finding suggests the decoupling is not tied to any particular image context. Rather, it appears to be a consistent phenomenon across medical domains. The results emphasize our analysis: ASR on its own does not adequately capture system vulnerability to adversarial attacks. 

\section{Discussion}\label{sec:disc}
This section explains that adversarial success behaves independently from perceptual quality and perturbation size, highlighting the need for multi-metric evaluation to better understand risks in medical AI systems.
\subsection{Decoupling of ASR from Perceptual and Distortion}
This study reveals a consistent misalignment between ASR and the other evaluation metrics. Whether we examine CNNs, ViTs, or both together, ASR shows
weak correlations with SSIM, PSNR, and $L_2$. In practice, this tells us that how large a perturbation is or how much it degrades image quality does not reliably
predict whether an attack will succeed. ASR appears to be picking up on something different, something more closely tied to classification outcomes than to the
perturbation itself. This is not to say that perceptual or distortion metrics are unimportant. What our results show is that these metrics and ASR capture
complementary sides of adversarial examples, each surfacing information the others do not. For example, two attacks that are against each other may have the same ASR values but very different perceptual qualities and perturbation magnitudes. ASR alone can't show these kinds of differences, which shows how limited single-metric evaluation is. The flatness of ASR across perturbation budgets (Figure~\ref{fig:asr_vs_eps}) offers further evidence: raising $\epsilon$ from 2/255 to 32/255 results in only slight ASR enhancements (e.g., CNN black-box: 75.5-79.0\%), even with a 16-fold rise in perturbation magnitude and a corresponding decline in SSIM and PSNR.

\subsection{Consistency Across Architectures}
Contrary to common assumptions that convolutional architectures exhibit more predictable adversarial behavior than transformer-based models, our results show that the weak ASR correlation persists in both CNNs and ViTs. While perceptual and distortion-based metrics maintain strong internal relationships in both architectures, ASR remains largely independent in all cases. This suggests the limitation is fundamental to how adversarial success is defined, rather than being an artifact of any particular model design.
However, the transferability analysis shows an important architectural distinction: ViTs show uniformly high cross-model transfer (91.1-93.4\% black-box ASR) compared to CNN's structured, asymmetric transferability (72-82.4\%). 

\subsection{Implications for Medical Image Analysis Systems}
In medical image analysis, the consequences of an undetected adversarial attack extend well beyond misclassification. As a result, the misclassification can translate directly into patient harm. Adversarial perturbations that preserve high perceptual fidelity, reflected in elevated SSIM and PSNR, while simultaneously achieving high ASR. The findings represent a concerning threat in clinical workflows. Interestingly, the perturbations may be visually indistinguishable from unperturbed images during review by radiologists or pathologists. In our study, we found that attacks introducing substantial and visually apparent distortions can provide ASR values on par with those of near-imperceptible perturbations. The consistently high ViT transferability, as shown in Figure ~\ref{fig:transfer_matrices}b, makes the problem even worse. We also found that an adversary who has access to any single ViT model can make attacks that transfer to other ViT models with over 91\% success, no matter how much money they have to spend on perturbations.

\section{Conclusion}\label{sec:con}

Our study evaluates adversarial transferability across four medical image datasets using seven models, seven attacks, and five perturbation budgets, providing 3,500 configurations. Our results reveal consistent decoupling
between ASR, perceptual, and distortion metrics. We found that the ASR correlates weakly with SSIM, PSNR, and $L_2$ across CNNs and ViTs ($|r| \leq 0.17$, $|\rho| \leq 0.15$), while these three metrics remain tightly interrelated
($|r| \geq 0.87$, $|\rho| \geq 0.97$). A 16-fold increase in perturbation budget yields only marginal ASR gains (CNN black-box: 75.5\% to 79.0\%) despite considerable quality degradation. We observe that transferability differs across architectures: ViTs transfer uniformly well (91.1\%-93.4\% black-box ASR), while CNN transferability varies, with Inception-v3 being the most transferable and ResNet50 the most resistant. 
The Ensemble attack narrows the white-box vs. black-box gap to 1.1\%, confirming gradient aggregation as the most effective strategy. We found that the CNN black-box ASR drops to 52.8\% while ViTs reach 100\%, showing multi-label and architectural interactions that any single metric captures. In this study, we show that ASR alone is insufficient for characterizing adversarial risk. We also jointly consider attack efficacy, perceptual degradation, and perturbation cost to find the attack success. Our future work will focus on a composite metric unifying ASR, perceptual quality, and perturbation magnitude. The metrics would provide a more holistic risk assessment than any single measure. 

\section{Data  Availability}
\label{sec:availability}
To support the reproducibility of our findings, the source code, datasets, and experimental artifacts used in this study are hosted on the Open Science Framework (OSF) and can be accessed at: \url{https://osf.io/4grsd/overview?view_only=c6246a2f9ea34a39a557da080c95ebed}.
{\footnotesize \bibliographystyle{ieeetr}
\bibliography{references}

\end{document}